\documentclass[runningheads,a4paper]{llncs}
\usepackage{amssymb, amsmath, bm, latexsym,comment}
\usepackage{multirow}
\usepackage{xcolor}
\usepackage{graphicx}
\usepackage{subfigure}
\usepackage{indentfirst}
\usepackage{setspace}
\usepackage{verbatim}
\usepackage{array}
\newcolumntype{L}[1]{>{\raggedright\let\newline\\\arraybackslash\hspace{0pt}}m{#1}}
\newcolumntype{C}[1]{>{\centering\let\newline\\\arraybackslash\hspace{0pt}}m{#1}}
\newcolumntype{R}[1]{>{\raggedleft\let\newline\\\arraybackslash\hspace{0pt}}m{#1}}

\providecommand{\zxhrefeq}[1]{(\ref{#1})}

\providecommand{\zxhreftb}[1]{Table~\ref{#1}}
\providecommand{\zxhreffig}[1]{Fig.~\ref{#1}}
\providecommand{\Zxhreffig}[1]{Fig.~\ref{#1}}

\providecommand{\citep}[1]{\cite{#1}}

\usepackage{url}
\urldef{\mailsa}\path|{alfred.hofmann, ursula.barth, ingrid.haas, frank.holzwarth,|
\urldef{\mailsb}\path|anna.kramer, leonie.kunz, christine.reiss, nicole.sator,|
\urldef{\mailsc}\path|erika.siebert-cole, peter.strasser, lncs}@springer.com|

\begin{document}

\mainmatter  
\title{Atrial fibrosis quantification based on maximum likelihood estimator of multivariate images}

\titlerunning{Atrial fibrosis quantification based on maximum likelihood estimator}

\author{Fuping Wu\inst{1} \and Lei Li \inst{2} \and
Guang Yang \inst{3} \and Tom Wong \inst{3} \and Raad Mohiaddin \inst{3} \and David Firmin \inst{3} \and Jennifer Keegan \inst{3} \and
Lingchao Xu \inst{2} \and
Xiahai Zhuang\inst{1}\thanks{Corresponding author: zxh@fudan.edu.cn. This work was supported by Science and Technology Commission of Shanghai Municipality (17JC1401600).}%
}
\authorrunning{Fuping Wu et al.}   
%
\tocauthor{}

\institute{School of Data Science, Fudan University, Shanghai, China
 \and
 School of BME and School of NAOCE, Shanghai Jiao Tong University
 \and
 National Heart and Lung Institute, Imperial College London, London, UK
 }

\maketitle

\begin{abstract}
We present a fully-automated segmentation and quantification of the left atrial (LA) fibrosis and scars combining two cardiac MRIs, one is the target late gadolinium-enhanced (LGE) image, and the other is an anatomical MRI from the same acquisition session. We formulate the joint distribution of images using a multivariate mixture model (MvMM), and employ the maximum likelihood estimator (MLE) for texture classification of the images simultaneously. The MvMM can also embed transformations assigned to the images to correct the misregistration. The iterated conditional mode algorithm is adopted for optimization. This method first extracts the anatomical shape of the LA, and then estimates a prior probability map. It projects the resulting segmentation onto the LA surface, for quantification and analysis of scarring. We applied the proposed method to 36 clinical data sets and obtained promising results (Accuracy: $0.809\pm .150$, Dice: $0.556\pm.187$). We compared the method with the conventional algorithms and showed an evidently and statistically better performance ($p<0.03$).
\end{abstract}

\section{Introduction}


Atrial fibrillation (AF) is the most common arrhythmia of clinical significance. It is associated with structural remodelling, including fibrotic changes in the left atrium (LA) and can increase morbidity.
Radio frequency ablation treatment aims to eliminate AF, which requires LA scar segmentation and quantification.
There are well-validated imaging methods for fibrosis detection and assessment in the myocardium of the ventricles such as the late gadolinium-enhanced (LGE) MRI. And recently there is a growing interest in imaging the thin LA walls for the identification of native fibrosis and ablation induced scarring of the AF patients \citep{journal/rad/Akcakaya12}.

Visualisation and quantification of atrial scarring require the segmentation from the LGE MRI images.
Essentially, there are two segmentations required: one showing the cardiac anatomy, particularly the LA and pulmonary veins, and the other delineating the scars.
The former segmentation is required to rule out confounding enhanced tissues from other substructures of the heart, while the latter is a prerequisite for analysis and quantification of the LA scarring.
While manual delineation can be subjective and labour-intensive, automating this segmentation is desired but remains challenging mainly due to two reasons.
First, the LA wall, including the scar, is  thin and sometimes hard to distinguish even by experienced cardiologists.
Second, the respiratory motion and varying heart rates can result in poor quality of the LGE MRI images.
Also, artifactually enhanced signal from surrounding tissues can confuse the algorithms.

\begin{figure*}[t]\center \begin{tabular}{@{} l  r @{}l@{}}
\includegraphics[height= 0.28\textwidth]{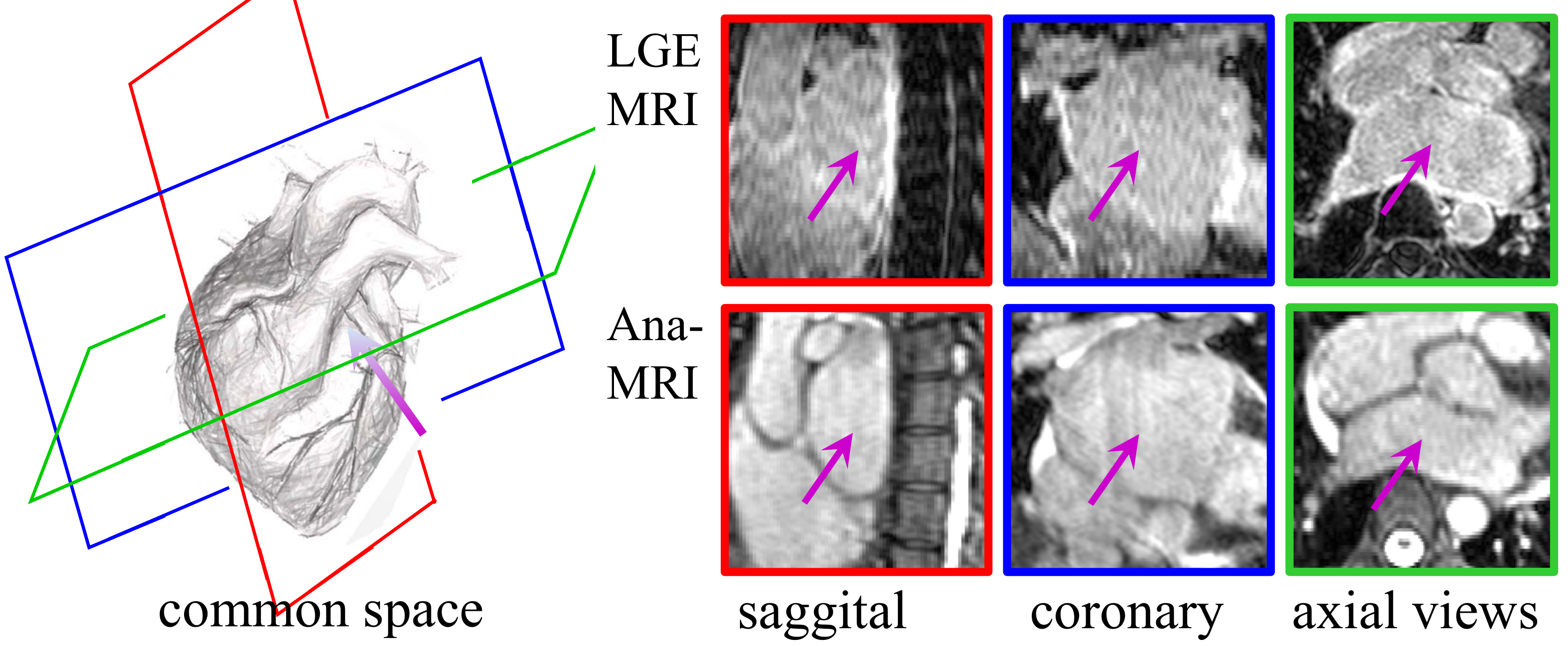} \quad &
 \raisebox{2ex}{\includegraphics[height= 0.25\textwidth]{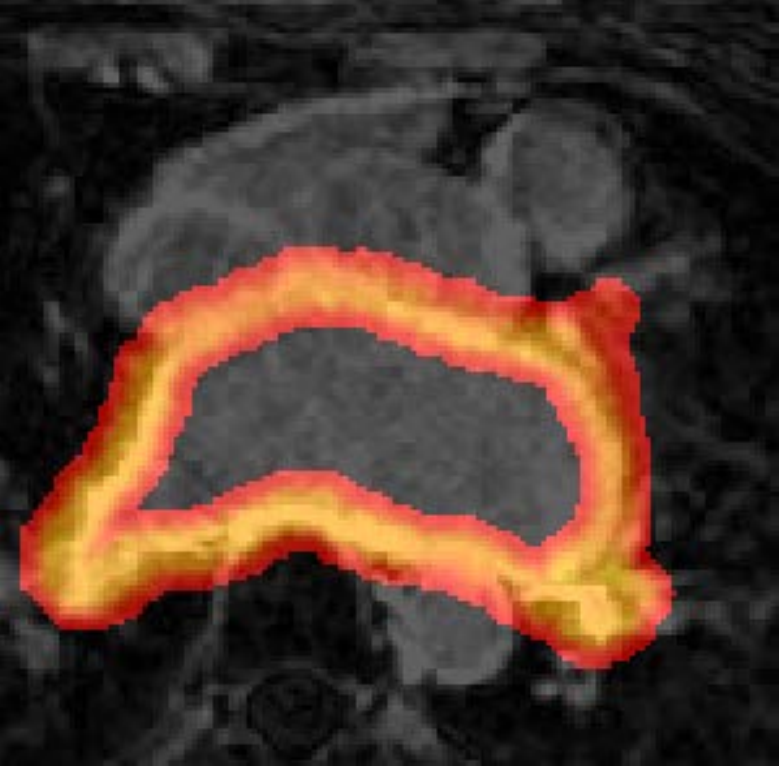}} &\!
 \raisebox{6.4ex}{\includegraphics[height= 0.20\textwidth]{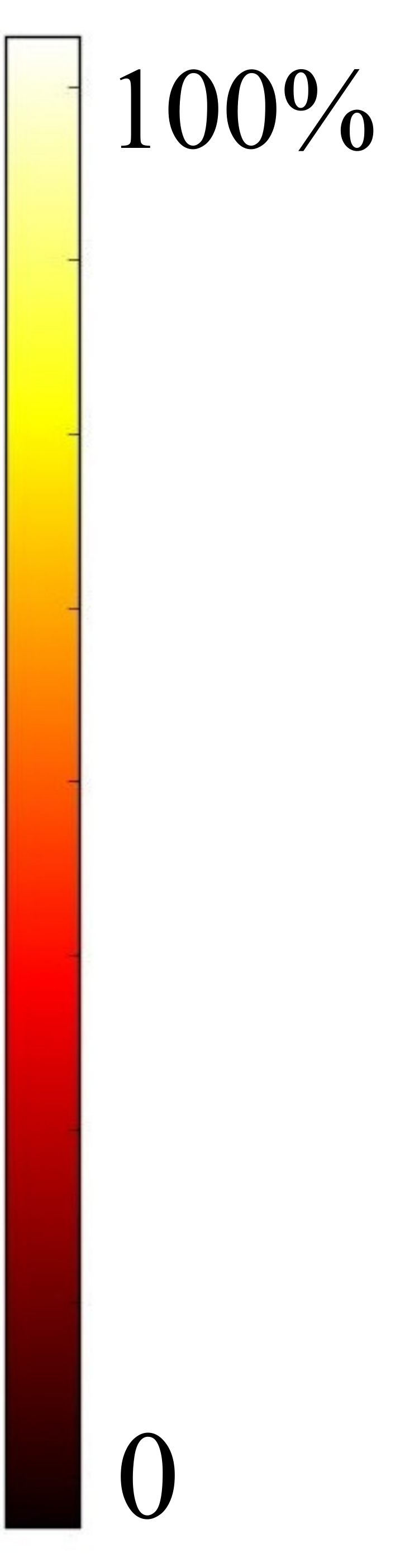}} \\[-2ex]
 \end{tabular}
   \caption{Illustration of the common space, MRI images, and LA wall probability map.}
\label{fig:images}\end{figure*}

Limited number of studies have been reported in the literature to develop fully automatic LA segmentation and quantification of scarring. Directly segmenting the scars has been the focus of a number of works \citep{journal/jcmr/Karim13},  which generally require delineation of the LA walls manually \citep{journal/cir/Oakes09,conf/spie/Perry12}, thus some researchers directly dedicated to the automated segmentation of the walls \citep{conf/isbi/Veni13,conf/isbi/Veni15}.
Tobon-Gomez et al. organized a grand challenge evaluating and benchmarking LA blood pool segmentation with promising outcomes \citep{journal/tmi/Tobon15}.
In MICCAI 2016, Karim et al. organized a LA wall challenge on 3D CT and T2 cardiac MRI \citep{conf/stacom/Karim17}.
Due to the difficulty of this segmentation task, only two of the three participants contributed to automatically segmenting the CT data, and no work on the MRI data has been reported.

In this study, we present a fully automated LA wall segmentation and scar delineation method combining two cardiac MRI modalities, one is the target LGE MRI, and the other is an anatomical 3D MRI, referred to as Ana-MRI, based on the  balanced-Steady State Free Precession (bSSFP) sequence, which provides clear whole heart structures.
The two images are aligned into a commons space, defined by the coordinate of the patient, as \zxhreffig{fig:images} illustrates.
Then, a multivariate mixture model (MvMM) and the maximum likelihood estimator (MLE) are used for label classification.
In addition, the MvMM can embed transformations assigned to each image,
and the transformations and model parameters can be optimized by the iterated conditional modes (ICM) algorithm within the framework of MLE.
In this framework, the clear anatomical information from the Ana-MRI provides a global guidance for the segmentation of the LGE MRI, and the enhanced LA scarring in the LGE MRI enables the segmentation of fibrosis which is invisible in Ana-MRI.

\begin{figure*}[t]\center
 \framebox{\includegraphics[width=0.95\textwidth]{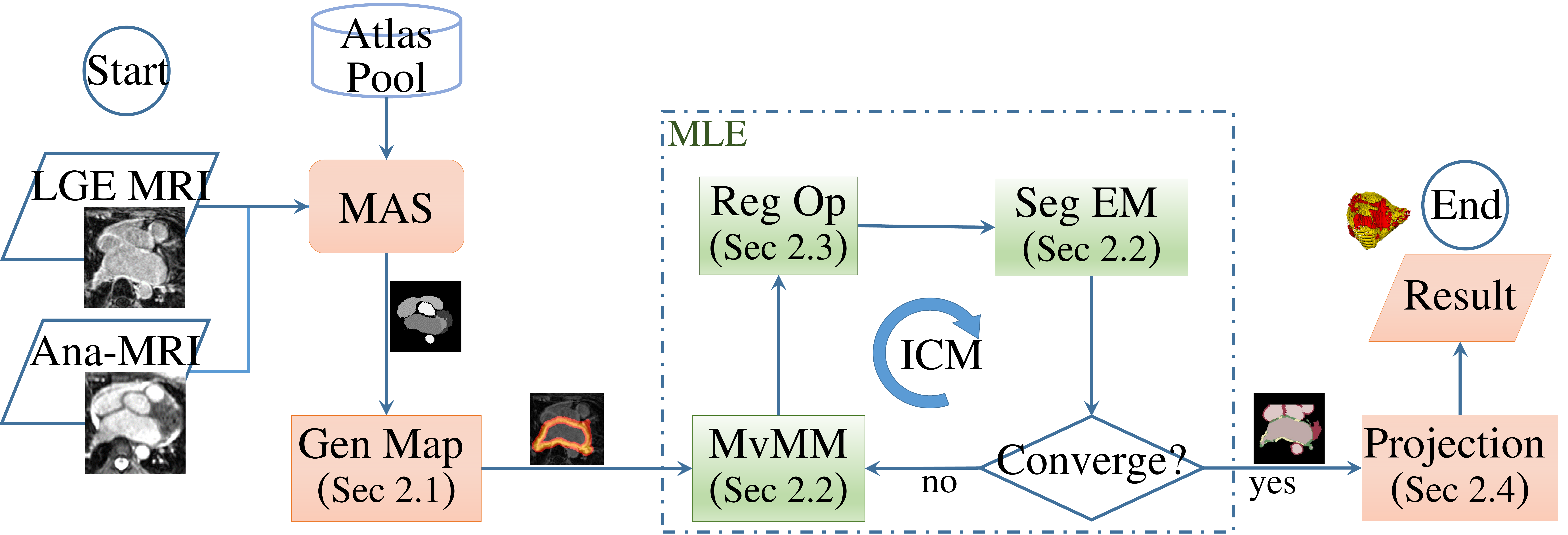}} \\[-1.2ex]
   \caption{Flowchart of the proposed LA wall segmentation combining two MRI sequences.}
\label{fig:flowchart}\end{figure*}

\section{Method} \label{method}
\vspace{-2pt}

The goal of this work is to obtain a fully automatic segmentation of LA wall and scars, combining the complementary information from the LGE MRI and Ana-MRI.
\Zxhreffig{fig:flowchart} presents the flowchart of the method, which includes three steps:
(1) A multi-atlas segmentation (MAS) approach is used to extract the anatomy of the LA, based on which the probability map is generated.
(2) The MLE and MvMM-based algorithm is performed to classify the labels and register the images.
(3) Projection of the resulting segmentation onto the LA surface for quantification and analysis.
\vspace{-2pt}

\subsection{MAS for generating LA anatomy and probability map}

The MAS consists of two major steps: (1) atlas propagation based on image registration and (2) label fusion.
Directly registering the atlases to the target LGE MRI can introduce large errors, as the LGE MRI has relative poor quality in general, and consequently results in inaccurate probability map for the thin LA walls. We therefore propose to use atlases constructed from a set of Ana-MRI images and register these Ana-MRI atlases to the target space of the subject, where both the Ana-MRI and LGE MRI have been acquired.
The inter-subject (atlas-to-patient) registration of Ana-MRI has been well developed \citep{journal/tmi/Zhuang10},
and the Ana-MRI to LGE MRI from the same subject can be reliably obtained by conventional registration techniques, since there only exists small misalignment between them.
For the label fusion, the challenge comes from the fact that the target LGE MRI and the Ana-MRI atlases have very different texture patterns, sometimes referred to as different-modality images even though they are both obtained from cardiac MRI. We hence use the multi-modality label fusion algorithm based on the conditional intensity of images \citep{journal/mia/Zhuang16}.

Having done the MAS, one can estimate a probability map of the LA wall by applying a Gaussian function, e.g. with zero mean and 2 mm standard deviation, to the boundary of the LA segmentation results assuming a fixed wall thickness for initialization \citep{journal/rad/Akcakaya12}. The probability of background can be computed by normalizing the two labels. \zxhreffig{fig:images} displays a slice of the LA wall probability map superimposed onto the LGE MRI.

\subsection{MvMM and MLE for multivariate image segmentation} \label{method:mvmm}

Let $\hat{I}\!\!=\!\!\{I_1\!\!=\!\!I_{LGE}, I_2\!\!=\!\!I_{Ana}\}$ be the two MRI images. We denote the spatial domain of the region of interest (ROI) of the subject as $\Omega$, referred to as \emph{the common space}.
\Zxhreffig{fig:images} demonstrates the common space and images.
For a location $x\in\Omega$, the label of $x$, i.e. LA wall or none LA wall (background), is determined regardless the appearance of the MRI images.
We denote the label of $x$ using $s(x)\!\!=\!\!k, k\in K$.
Provided that the two images are both aligned to the common space, the label information of them should be the same.
For the LA wall in LGE MRI, the intensity values are distinctly different for the fibrosis and normal myocardium.
We denote the subtype of a tissue $k$ in image $I_i$ as $z_i(x)\!\!=\!\!c$, $c\!\in\!\! C_{ik}$ and use the multi-component Gaussian mixture to model the intensity of the LA walls in LGE MRI.


The likelihood ($LH$) of the model parameters $\theta$ in MvMM is given by $LH(\theta;\hat{I})$ = $p(\hat{I}|\theta)$, similar to the conventional Gaussian mixture model (GMM) \citep{journal/tmi/LeemputMS99}.
Assuming independence of the locations (pixels), one has $LH(\theta;\hat{I})$=$\prod_{x\in\Omega}p(\hat{I}(x)|\theta)$.
In the EM framework, the label and component information are considered as hidden data.
Let $\Theta$ denotes the set of both hidden data and model parameters,
the likelihood of the complete data is then given by,\\[-1ex]
\begin{equation}
p(\hat{I}(x)|\Theta)=\sum_{k\in K} \pi_{kx} p(\hat{I}(x)| s(x)\!\!=\!\!k,\Theta), \\[-1ex]
\end{equation}
where, $\pi_{kx}\!\!=\!\!p(s(x)\!\!=\!\!k|\Theta)\!\!=\!\!\frac{p_A(s(x)=k) \pi_k}{N\!\!F}$, $p_A(s(x)\!\!=\!\!k)$ is the prior probability map, $\pi_k$ is the label proportion, and $N\!\!F$ is the normalization factor.

When the tissue type of a position is known, \emph{the intensity values from different images then become  independent},\\[-1ex]
\begin{equation}
p(\hat{I}(x)|s(x)\!\!=\!\!k,\Theta)=\prod_{i=1,2}  p(I_i(x)| s(x)\!\!=\!\!k,\Theta) \ .\\[-1ex]
\label{eq:labelintensity}
\end{equation}
Here, the intensity PDF of an image is given by the conventional GMM.
To estimate the Gaussian model parameters and then segmentation variables, one can employ the EM to solve the log-likelihood (LL) by rewriting it as follows,\\[-1ex]
\begin{equation}
 LL=\sum_x\sum_k \delta_{s(x),k}\Big( \log\pi_{kx} +\sum_i\sum_{c_{ik}} \delta_{z_i(x),c_{ik}}(\log \tau_{ikc}+\log \Phi_{ikc}(I_i(x))) \Big) ,\\[-1ex]
\label{eq:ll}
\end{equation}
where $\delta_{a,b}$ is the Kronecker delta function, $\tau_{ikc}$ is the component proportion and $\Phi_{ikc}(\cdot)$ is the Gaussian function to model the intensity PDF of a tissue subtype $c$ belonging to a tissue $k$ in the image $I_i$.
The model parameters and segmentation variables can be estimated using the EM algorithm and related derivation.
Readers are referred to the supplementary materials for details of the derivation.


\subsection{Optimization strategy for registration in MvMM}

The proposed MvMM can embed transformations for the images ($\{F_i\}$) and map ($F_m$), such as
$p(I_i(x)|c_{ik},\theta,F_i)\!=\!\Phi_{ikc}(I_i(F_i(x)))$, and
$p_A(s(x)\!\!=\!\!k|F_a) \!= \!A_k(F_a(x)),k=[l_{bk},l_{la}]$,
where $\{A_k(\cdot)\}$ are the probabilistic atlas image.
With the deformation embedded prior $\pi_{kx|F_m}\!\!=\!\!p(s(x)\!\!=\!\!k|F_m)$,
the LL becomes,
\begin{equation}
  LL
= \displaystyle\sum_{x\in{\Omega}} \log LH(x)
= \displaystyle\sum_{x\in{\Omega}} \log
        \Big\{ \sum_k \pi_{kx|F_m}
             \prod_i \sum_{c_{ik}} \tau_{ikc}\Phi_{ikc}(I_i(F_i(x))) \Big\}.
\label{eq:mvill}\end{equation}
Here, the short form $LH(x)$ is introduced for convenience.

There is no closed form solution for the minimization of \zxhrefeq{eq:mvill}.
Since the Gaussian parameters depend on the values of the transformation parameters, and vice versa, one can use the ICM approach to solve this optimization problem,
which optimizes one group of parameters while keeping the others unchanged at each iteration. The two groups of parameters are alternately optimized and this alternation process iterates until a local optimum is found.
The MvMM parameters and the hidden data are updated using the EM approach, and the transformations are optimized using the gradient ascent method.
The derivatives of LL with respect to the transformations of the MRI images and probability map are respectively given by,
\begin{equation} \begin{array}{@{}l@{}}
\displaystyle\frac{\partial LL}{\partial F_i} \!=\! \displaystyle\sum_x \frac{1}{LH(x)}\sum_k\!\pi_{kx}\! \prod_{j\neq i} \Big\{ p(I_j(x)|k_x,\theta, F_j) \sum_c \!\tau_{ikc}\Phi'_{ikc}\nabla I_i(y)\!\!\times\!\!\nabla F_i(x) \Big\}   \\
\mathrm{and\ }
\displaystyle\frac{\partial LL}{\partial F_m} \!=\! \displaystyle\sum_x \frac{1}{LH(x)}\sum_k  \frac{\partial \pi_{kx|F_m}}{\partial F_m}
 p(\boldsymbol{I}(x)|k_x,\theta,\{F_i\}) \ , \\[-2ex]\end{array}
\end{equation}
where $y\!\!=\!\!F_{i}(x)$.
The computation of $\frac{\partial \pi_{kx|F_m}}{\partial F_m}$ is related to $\frac{\partial A_k(F_m(x))}{\partial F_m}$, which equals $\nabla A_k(F_m(x))\!\!\times\!\! \nabla F_m(x)$.
Both $F_m$ and $\{F_i\}$ are based on the free-form deformation (FFD) model concatenated with an affine transformation,
which can be denoted as $F \!\!=\!\! G(D(x))$, where  $G$ and $D$ are respectively the affine and FFD transformations \citep{journal/TMI/Rueckert99}.

\subsection{Projection of the segmentation onto the LA surface}
The fibrosis is commonly visualized and quantified on the surface of the LA, focusing on the area and position of the scarring similar to the usage of EAM system \citep{journal/jice/Williams17}. Following the clinical routines, we project the classification result of scarring onto the LA surface extracted from the MAS, based on which the quantitative analysis is performed.

\section{Experiments}

\subsection{Materials}

\textbf{Data Acquisition:}
Cardiac MR data were acquired on a Siemens Magnetom Avanto 1.5T scanner (Siemens Medical Systems, Erlangen, Germany).
Data were acquired during free-breathing using a crossed-pairs navigator positioned over the dome of the right hemi-diaphragm with navigator acceptance window size of 5 mm and CLAWS respiratory motion control \citep{journal/mrm/Keegan14}.
The LGE MRI were acquired with resolution $1.5\!\times\!1.5\!\times\!4$ mm and reconstructed to $.75\!\times\!.75\!\times\!2$ mm, the Ana-MRI were acquired with $1.6\!\times\!1.6\!\times\!3.2$ mm, and reconstructed to $0.8\!\times\!0.8\!\times\!1.6$ mm.
\Zxhreffig{fig:images} provides an example of the images within the ROI.

\textbf{Patient information:}
In agreement with the local regional ethics committee, cardiac MRI was performed in longstanding persistent AF patients.
Thirty-six cases had been retrospectively entered into this study. 

\textbf{Ground truth and evaluation:}
The 36 LGE-MRI images were all manually segmented by experienced radiologists specialized in cardiac MRI to label the enhanced atrial scarring regions, which were considered as the ground truth for evaluation of the automatic methods.
Since, the clinical quantification of the LA fibrosis is made with the EAM system which only focuses on the surface area of atrial fibrosis,
both the manual and automatic segmentation results were projected onto the LA surface mesh \citep{journal/jice/Williams17}.
The Dice score of the two areas in the projected surface was then computed as the accuracy of the scar quantification.
The Accuracy, Sensitivity and Specificity measurements between the two classification results were also evaluated.

\textbf{Atlases for MAS and probability map:}
First we obtained 30 Ana-MRI images from the KCL LA segmentation grand challenge, together with manual segmentations of the left atrium, pulmonary veins and appendages \citep{journal/tmi/Tobon15}. In these data, we further labelled the left and right ventricles, the right atrium, the aorta and the pulmonary artery, to generate 30 whole heart atlases for target-to-image registration. These 30 images were employed only for building an independent multi-atlas data set, which will then be used for registering to the Ana-MRI data that linked with the LGE MRI scans of the AF patients.

\subsection{Result}

\begin{table}[t]\center
\caption{Quantitative evaluation results of the five schemes.}
\begin{tabular}{@{\ }l@{\ }|@{\ }c@{\ }c@{\ }c@{\ }c@{\ }} \hline
Method & Accuracy & Sensitivity & Specificity & Dice \\\hline
OSTU$^{+\texttt{AnaMRI}}$ &
$0.395\pm0.181$&$0.731\pm0.165$&$0212\pm0.115$&$0.281\pm0.129$\\
GMM$^{+\texttt{AnaMRI}}$ &
$0.569\pm0.132$&$0.950\pm0.164$&$0.347\pm0.133$&$0.464\pm0.133$\\
MvMM &
$0.809\pm0.150$&$0.905\pm0.080$&$0.698\pm0.238$&$0.556\pm0.187$\\
\hline
\end{tabular}\label{tb:result}\end{table}

\begin{figure*}[t]\center \begin{tabular}{@{}l   r @{}}
\includegraphics[height=0.42\textwidth,width=0.44\textwidth]{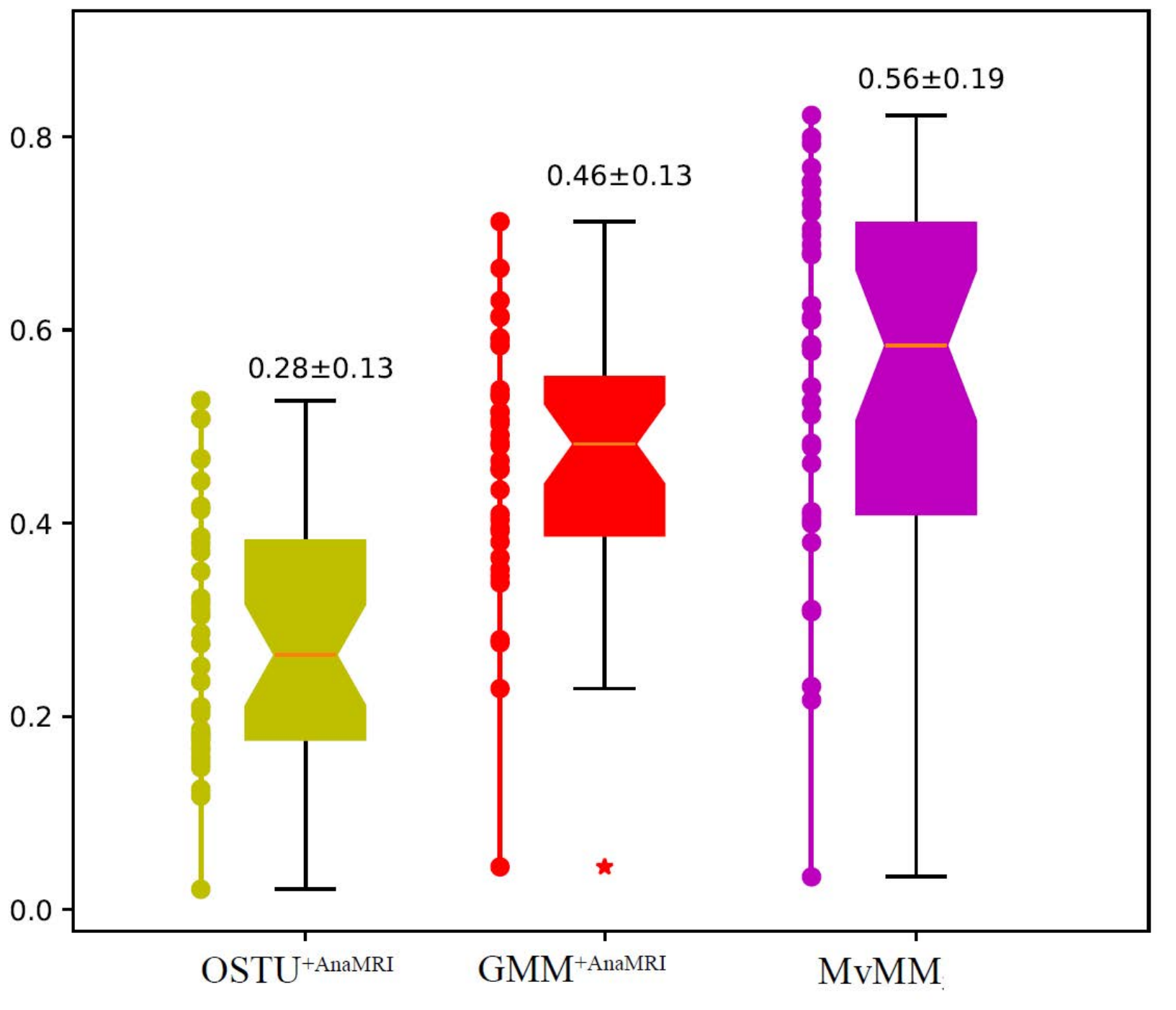}&
\raisebox{-0.3ex}{\includegraphics[height= 0.42\textwidth ]{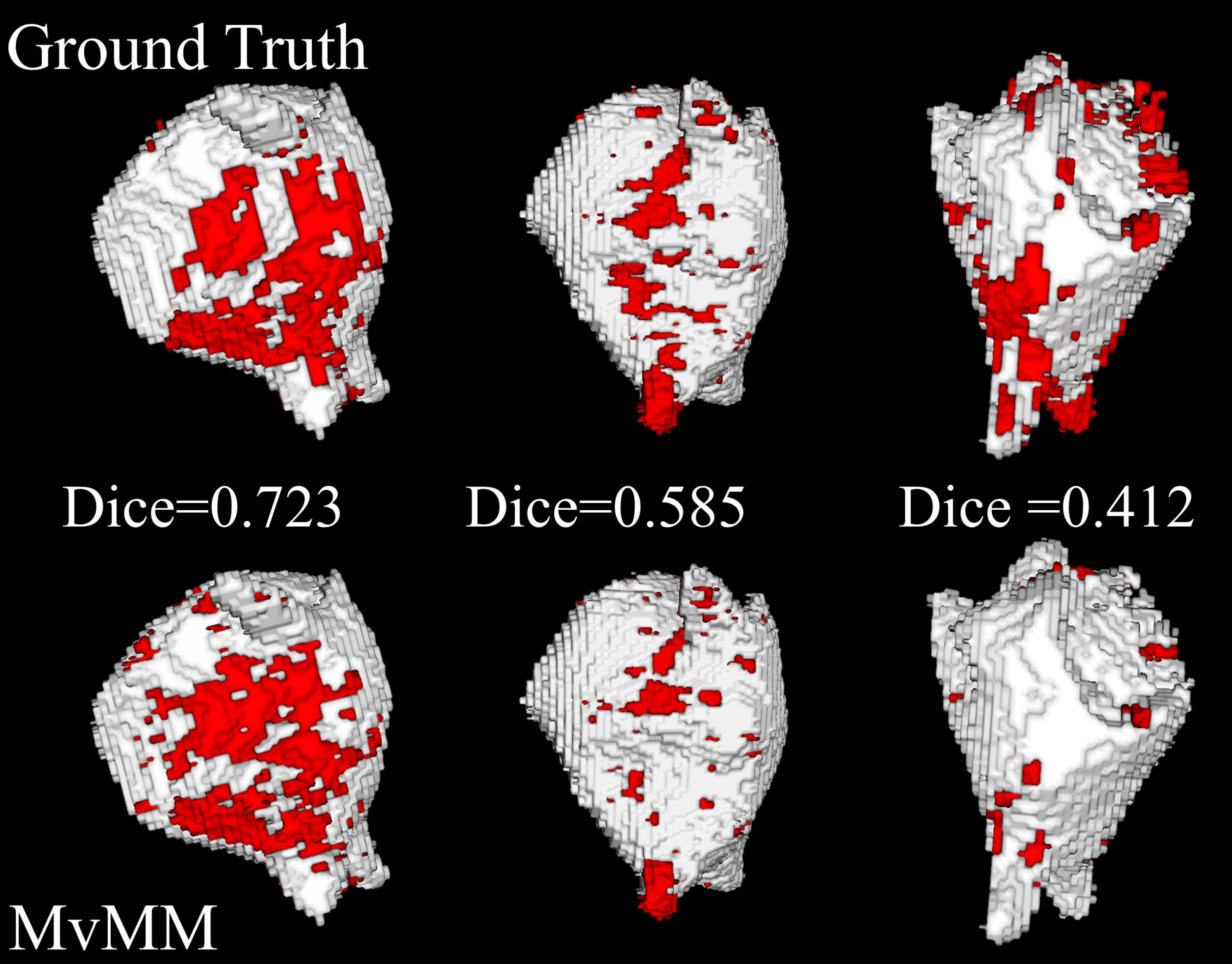}}\\[-2ex]
 \end{tabular}
\caption{Left: the box plots of the three results. Right: the 3D visualization of the three cases from the first quarter, median and third quarter of MvMM segmentation in terms of Dice with the ground truth.}
\label{fig:result}\end{figure*}

For comparisons, we included the results using OSTU threshold \citep{journal/ostu} and conventional GMM \citep{journal/tmi/LeemputMS99}.
Both the two schemes however could not generate scar segmentation directly from the LGE MRI.
Therefore, we employed the whole heart segmentation results from combination of Ana-MRI and LGE MRI. We generated a mask of the LA wall for OSTU threshold and used the same probability map of LA wall for GMM.
Therefore, the two methods are indicated as OSTU$^{+\texttt{AnaMRI}}$ and GMM$^{+\texttt{AnaMRI}}$, respectively.

\zxhreftb{tb:result} presents the quantitative statistical results of three methods, and \zxhreffig{fig:result} (left) provides the corresponding box plots. Here, the proposed method is denoted as MvMM.
The proposed method performed evidently better than the two compared methods with statistical significance ($p<0.03$), even though both OSTU and GMM used the initial segmentation of LA wall or probabilistic map computed from MAS of the combined LGE MRI and Ana-MRI. It should be noted that without Ana-MRI, the direct segmentation of the LA wall from LGE MRI could fail, which results in a failure of the LA scar segmentation or quantification by the OSTU or GMM.

\Zxhreffig{fig:result} (right) visualizes three examples for illustrating the segmentation and quantification of scarring for clinical usages.
These three cases were selected from the first quarter, median, third quarter cases of the test subjects according to their Dice scores by the proposed MvMM method.
The figure presents both the results from the manual delineation and the automatic segmentation by MvMM for comparisons.
Even though the first quarter case has much better Dice score, the accuracy of the localizing and quantifying the scarring can be similar to that of the other two cases.
This is confirmed by the comparable results using the other measurements as indicators of quantification performance,
0.932 VS 0.962 VS 0.805 (Accuracy),
0.960 VS 0.973 VS 0.794 (Sensitivity) and
0.746 VS 0.772 VS 0.983 (Specificity) for the three cases.
This is because when the scarring area is small, the Dice score of the results tends to be low.
Note that for all the pre-ablation scans of our AF patients, the scars may be relatively rare to see.

\section{Conclusion}
We have presented a new method, based on the maximum likelihood estimator of multivariate images, for LA wall segmentation and scar quantification, combining the complementary information of two cardiac MRI modalities.
The two images of the same subject are aligned to a common space and the segmentation of them is performed simultaneously.
To compensate the deformations of the images to the common space, we formulate the MvMM with transformations and propose to use ICM to optimize the different groups of parameters.
We evaluated the proposed techniques using 36 data sets acquired from AF patients.
The combined segmentation and quantification of LA scarring yielded promising results, Accuracy: 0.809, Sensitivity: 0.905, Specificity: 0.698, Dice: 0.556, which is difficult to achieve for the methods solely based on single-sequence cardiac MRI.
In conclusion, the proposed MvMM is a generic, novel and useful model for multivariate image analysis. It has the potential of achieving good performance in other applications where multiple images from the same subject are available for complementary and simultaneous segmentation.


\bibliographystyle{splncs}

\bibliography{../AllBibliography20160615} 

\end{document}